\pdfoutput=1

\documentclass[11pt]{article}

\usepackage[preprint]{acl}

\usepackage{times}
\usepackage{latexsym}

\usepackage[T1]{fontenc}

\usepackage[utf8]{inputenc}

\usepackage{microtype}

\usepackage{inconsolata}

\usepackage{graphicx}
\usepackage{amsmath,amsfonts,amssymb}
\usepackage{array}
\usepackage[caption=false,font=normalsize,labelfont=sf,textfont=sf]{subfig}
\usepackage{textcomp}
\usepackage{stfloats}
\usepackage{url}
\usepackage{verbatim}
\usepackage{graphicx}
\usepackage[affil-it]{authblk}
\usepackage{balance}

\usepackage{tikz}
\usepackage[edges]{forest} 
\definecolor{hidden-draw}{RGB}{106,142,189} 
\definecolor{hidden-blue}{RGB}{194,232,247} 
\definecolor{hidden-orange}{RGB}{217, 232, 252} 
\definecolor{trainingColor}{RGB}{180, 0, 0} 
\definecolor{finetuningColor}{RGB}{0, 102, 204} 
\definecolor{noFinetuningColor}{RGB}{0, 128, 0} 
\definecolor{knowledgeColor}{RGB}{128, 0, 128} 
\definecolor{peftColor}{RGB}{255, 140, 0}

\usepackage{caption}

\usepackage{booktabs}
\usepackage{mathabx}
\usepackage{algorithm}
\usepackage[noend]{algpseudocode}
\usepackage{multirow}
\usepackage{wrapfig}

\usepackage{hyperref}

\title{Rethinking the Outlier Distribution in Large Language Models: An In-depth Study}

\author{
\textbf{Rahul Raman\textsuperscript{1}} \quad
\textbf{Khushi Sharma\textsuperscript{1}} \quad
\textbf{Sai Qian Zhang\textsuperscript{1}} \\
\textsuperscript{1}New York University \\
\texttt{\{rr4549, ks7406, sai.zhang\}@nyu.edu}
}

\begin{document}
\maketitle

\begin{abstract}
Investigating outliers in large language models (LLMs) is crucial due to their significant impact on various aspects of LLM performance, including quantization and compression. Outliers often cause considerable quantization errors, leading to degraded model performance. Identifying and addressing these outliers can enhance the accuracy and efficiency of the quantization process, enabling smoother deployment on edge devices or specialized hardware. Recent studies have identified two common types of outliers in LLMs: massive activations and channel-wise outliers. While numerous quantization algorithms have been proposed to mitigate their effects and maintain satisfactory accuracy, few have thoroughly explored the root causes of these outliers in depth.

In this paper, we conduct a comprehensive investigation into the formation mechanisms of these outliers and propose potential strategies to mitigate their occurrence. Ultimately, we introduce some efficient approaches to eliminate most massive activations and channel-wise outliers with minimal impact on accuracy.
\end{abstract}

\section{Introduction}
Large Language Models (LLMs) have emerged as a cornerstone in the field of natural language processing (NLP), transforming how we approach various linguistic tasks. These models, with their ability to understand and generate human-like text, have revolutionized applications ranging from conventional NLP tasks such as machine translation~\cite{huang2023towards, xu2024contrastive, zhu2023multilingual}, sentiment analysis~\cite{miah2024multimodal, wang2024llm, deng2023llms} to advanced tasks such as code generation~\cite{kazemitabaar2023novices, thakur2024verigen, nakkab2024rome}. However, the enormous size of LLMs, often reaching billions of parameters, presents substantial challenges for deployment, necessitating the use of techniques that enable efficient inference. 

To address this, Post-Training Quantization (PTQ)~\cite{frantar2022gptq, xiao2023smoothquant, lin2024awq, yao2022zeroquant} provides a practical, low-cost approach for model quantization, either completely training-free or with minimal calibration effort~\cite{cai2020zeroq,li2021brecq}. In comparison to Quantization-Aware Training (QAT), which demands multiple fine-tuning iterations, PTQ incurs much lower computational costs, making it suitable for LLM.
Unfortunately, outliers in LLM activations and KV vectors~\cite{dettmers2022gpt3, zeng2022glm} introduce significant magnitude variations among LLM elements, which in turn lead to a notable drop in model accuracy when low-precision PTQ is applied~\cite{xiao2023smoothquant, quip, ashkboos2024quarot}.

\begin{figure}[t]
    \centering
    \includegraphics[width=\columnwidth]{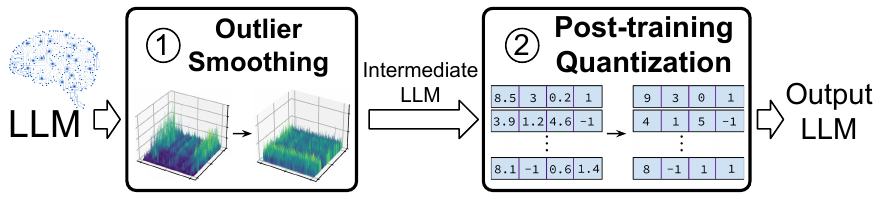}
    \caption{Given a pretrained LLM, techniques are first applied to smooth out the outliers in its activations. The resulting model is then quantized, achieving superior accuracy.}
    \label{fig:llm-step}
\end{figure}

Prior research has identified two types of outliers in LLM activations. The first, massive activations (MAs), commonly appear across various LLMs and are typically linked to specific tokens in certain channels~\cite{sun2024massive}. The second type, channel-wise outliers~\cite{dettmers2022gpt3, xiao2023smoothquant, ashkboos2024quarot}, manifests in bulk within specific channels. These findings have inspired a two-stage approach in modern quantization techniques: initially, methods are employed to eliminate outliers in the pretrained LLM, resulting in a model with a smoother value distribution in its activations. Subsequently, quantization algorithms such as GPTQ~\cite{frantar2022gptq} and OBQ~\cite{frantar2022optimal} are applied to produce low-precision LLMs, as shown in Figure~\ref{fig:llm-step}.

Outlier smoothing is a crucial step in achieving efficient LLM quantization. Understanding the root causes of outliers is essential for developing effective quantization techniques and gaining deeper insights into model behavior and robustness. While prior studies have identified the presence of MAs and channel-wise outliers, and proposed methods to mitigate them~\cite{sun2024massive, liu2024spinquant, bini2024characterizing, xiong2024uncomp}, none have explored the fundamental reasons behind the existence of these outliers from a numerical perspective, particularly with operator-level granularity. This finer-grained understanding is crucial, as different layers and operators may contribute uniquely to the formation and propagation of LLM outliers, influencing both performance and accuracy in low-precision LLMs.

\begin{figure*}[t]
    \centering
    \includegraphics[width=2.1\columnwidth]{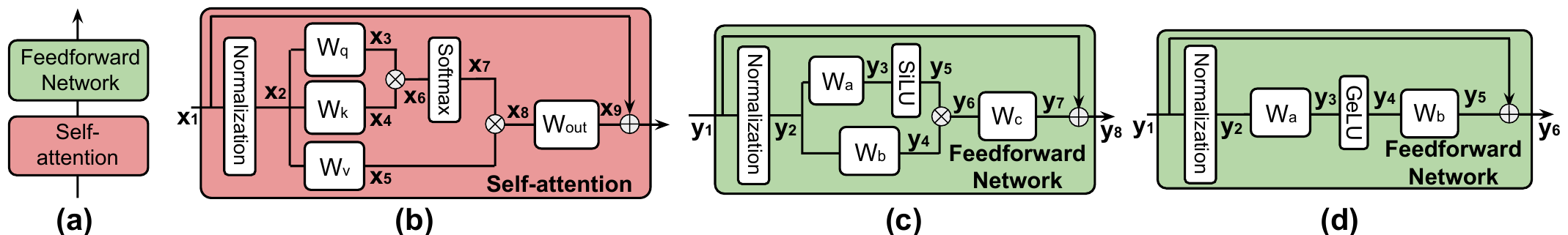}
    \caption{(a) Architecture of a LLM decoder block. (b), (c) and (d) show the architectures of self-attention block, standard FFN (conventional MLP), and gated FFN (GLU), respectively. The notations will be used throughout the rest sections.}
    \label{fig:llm-arch}
\end{figure*}
In this work, we investigate the underlying reasons for the existence of outliers in LLMs at the operator level through extensive empirical analysis. Our study provides valuable insights to guide the development of effective outlier smoothing algorithms. Building on these findings, we propose some novel methods to efficiently mitigate the majority of massive activations and channel-wise outliers without compromising model accuracy. This significantly reduces the complexity of subsequent LLM quantization processes. In summary, our findings on LLM outliers can be summarized as follows:

\begin{itemize}
    \item We empirically demonstrate that massive activations (MAs) are predominantly generated in the initial layers the model. Once these MAs arise, they persist throughout the LLM, being propagated through subsequent layers via residual connections.
    \item Previous studies indicate that the removal of MAs can significantly impact the quantization process. Surprising, our empirical analysis shows that eliminating MAs introduced by residual connections has no measurable effect on the model's accuracy. Notably, these MAs constitute the majority of MAs in LLMs.
    \item Channel-wise outliers in LLMs initially emerge due to the normalization operations within the model. The rescaling operation within the normalization layer exacerbates this issue by introducing an increasing number of channel-wise outliers.
    \item Certain channels within the weight matrices can also contribute to the emergence of channel-wise outliers in the intermediate results of LLMs.
\end{itemize}

\section{Background and Related Work}
\label{sec:related-work}
\subsection{LLM Operations}
\label{sec:bg:llm-arch}
Modern LLMs (e.g., Llama series~\cite{touvron2022llama,touvron2023llama}, GPT series~\cite{radford2019language,brown2020language}) are constructed as a stack of transformer decoders, with each decoder comprising two fundamental components: a Self-Attention (SA) block and a feed forward network (FFN), as depicted in Figure~\ref{fig:llm-arch} (a). During the LLM serving process, the input to the Self-Attention (SA) block is first processed by a normalization operation (e.g., LayerNorm or RMSNorm). As detailed in Figure~\ref{fig:llm-profile}(d), this normalization consists of two key steps: standardization and rescaling. Specifically, the input \(X\) is normalized by subtracting its mean \(\mu_X\) and dividing by its standard deviation \(\sigma_X\). Subsequently, each channel of the standardized output is scaled by a learnable parameter \(\gamma\) and shifted by another learnable parameter \(\beta\).

The output of the normalization operation is then multiplied with three weight matrices $W_{Q}$, $W_{K}$, and $W_{V}$, yielding the outputs referred to as query ($q$), key ($k$), and value ($v$), which is shown as $x_{3}$, $x_{4}$ and $x_{5}$ in Figure~\ref{fig:llm-arch}, respectively. 
The resulting $q$ and $k$, in combination with $v$, will then undergo multiplication, Softmax, and residual addition to generate the SA output, as shown in Figure~\ref{fig:llm-arch} (b).

The output from the SA will then be passed to the FFN for further processing, which typically involves a gated MLP~\cite{radford2018improving, radford2019language} (Figure~\ref{fig:llm-arch} (c)) or standard MLP~\cite{liu2021pay, touvron2022llama, touvron2023llama} (Figure~\ref{fig:llm-arch} (d)). The FFN consists of a normalization operation, multiple fully connected (FC) layers along with an intermediate activation function, such as GeLU~\cite{hendrycks2016gaussian} or SiLU~\cite{hendrycks2016gaussian}.

\subsection{Outlier in LLM}
\label{sec:bg:llm-outlier}
\begin{figure}
    \centering
    \includegraphics[width=\columnwidth]{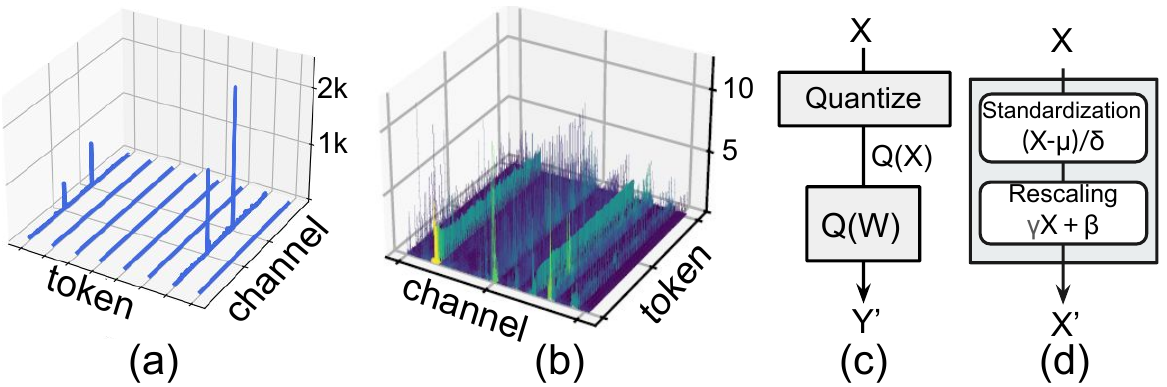}
    \caption{(a) One example of massive activation presented in the inputs $x_{1}$. (b) An example of outlier channel at position $x_{2}$ in the LLM. (c) The existence of outlier will lead to an output Y' different from the original output Y. (d) The normalization operations within LLM.}
    \label{fig:llm-profile}
\end{figure}
As prior studies have demonstrated~\cite{dettmers2022gpt3, zeng2022glm, sun2024massive}, outliers can be categorized into two types:~\textit{massive activations (MA)} and~\textit{channel-wise outliers (CO)}. The presence of outliers in LLM activations and KV vectors~\cite{dettmers2022gpt3, zeng2022glm} often causes a significant drop in model accuracy when low-precision PTQ is applied~\cite{xiao2023smoothquant, quip, ashkboos2024quarot}. 

While earlier research, such as~\cite{bondarenko2023quantizabletransformersremovingoutliers}, has shown that the attention mechanism can lead to excessive activations by concentrating too much on specific tokens, resulting in scenarios where the mechanism fails to remain inactive and creates an outlier problem, these studies mainly focus on BERT architectures. In contrast, our analysis expands the scope to include LLaMA, GPT, and Qwen models. This broader investigation provides new insights into the architectural changes that can give rise to activation outliers. As discussed in \cite{li2024evaluatingquantizedlargelanguage}, performing kurtosis on the activation tensor to reflect on MAs in GLUs, though our focus here remains on characterizing and categorizing the outlier phenomenon and providing a simpler method to remove MAs.

To demonstrate this, we profile the inputs \( x_{2} \) to the \( W_{Q} \), \( W_{K} \), and \( W_{V} \) matrices within the self-attention (SA) block, as shown in Figure~\ref{fig:llm-arch}, using the Wikitext dataset~\cite{wikitext} on the LLaMA-7B model. Following the notations in Figure~\ref{fig:llm-arch}, we record the input to the normalization operation, \( x_{1} \).
The results presented in Figure~\ref{fig:llm-profile} (a) highlights the presence of MAs in \(x_{1}\), with magnitudes often reaching thousands. Furthermore, these MAs propagate through the normalization operation, causing \(x_{2}\) to also exhibit some outliers. Although the magnitude of these outliers is reduced after normalization, they remain significant.
Additionally, Figure~\ref{fig:llm-profile} (b) shows that distribution of the COs in $x_{2}$, corroborating earlier findings~\cite{xiao2023smoothquant, ashkboos2024quarot, ashkboos2024slicegpt, frantar2023sparsegpt}. To isolate the impact of MAs, we remove them from \( x_{2} \) to better illustrate the distribution of COs.

Figure~\ref{fig:llm-profile} (c) illustrates that when the input \( X \) contains both types of outliers, its quantized version \( Q(X) \) experiences significant quantization error. As a result, the output \( Y' \), derived from the quantized input \( Q(X) \) and quantized weight \( Q(W) \), deviates considerably from the original output \( Y = XW \), leading to a noticeable degradation in accuracy.

\begin{table*}
    \centering
    \caption{TMA values distribution within different LLMs. Initial Top-1 and Initial Top-2 denote the MAs with the largest and second largest magnitudes within the initial LLM layers. Last-1 and Last-2 denote the last and second last layers within LLM. The N/A for smaller models indicate that the balancing of signs observed in True Massive Activations (TMAs) is handled only by the last layer. }
     \resizebox{\textwidth}{!}{
    \begin{tabular}{l>{\centering\arraybackslash}p{1.6cm} >{\centering\arraybackslash}p{1.6cm} | >{\centering\arraybackslash}p{1.6cm} >{\centering\arraybackslash}p{1.6cm} | >{\centering\arraybackslash}p{1.6cm} >{\centering\arraybackslash}p{1.6cm} | >{\centering\arraybackslash}p{1.6cm} >{\centering\arraybackslash}p{1.6cm} | >{\centering\arraybackslash}p{1.6cm} >{\centering\arraybackslash}p{1.6cm}}
        \toprule
        & \multicolumn{2}{c}{\textbf{LLaMA3.2-3B}} & \multicolumn{2}{c}{\textbf{LLaMA3.1-8B}} & \multicolumn{2}{c}{\textbf{LLaMA2-13B}} & \multicolumn{2}{c}{\textbf{GPT-2}} & \multicolumn{2}{c}{\textbf{Qwen2.5-7B}} \\
        \cmidrule(lr){2-3} \cmidrule(lr){4-5} \cmidrule(lr){6-7} \cmidrule(lr){8-9} \cmidrule(lr){10-11}
        \textbf{Massive Activations} & \textbf{Value} & \textbf{Position} & \textbf{Value} & \textbf{Position} & \textbf{Value} & \textbf{Position} & \textbf{Value} & \textbf{Position} & \textbf{Value} & \textbf{Position} \\
        \midrule
        \textit{Initial Top-1} & -328.25 & (0, 588) & -300.5 & (0, 788) & -1211.0 & (0, 4743) & -449.82 & (0, 1591) & -9057.43 & (0, 458)\\
        \textit{Initial Top-2} & -303.25 & (0, 1016) & -274.75 & (0, 1384) & -708.0 & (0, 2100) & -388.98 & (0, 506) & -5757.42 & (0, 2570) \\ \hline
        \textit{Last-2 Top-1} & N/A & N/A & N/A & N/A & 414.75 & (0, 4743) & 169.89 & (0, 1591) & 9178.38 & (0, 458) \\
        \textit{Last-2 Top-2} & N/A & N/A & N/A & N/A & 288.25 & (0, 2100) & 159.61 & (0, 506) & 4645.87 & (0, 2570) \\ \hline
        \textit{Last-1 Top-1} & 262.5 & (0, 1016) & 299.75 & (0, 788) & 824.0 & (0, 4743) & 277.06 & (0, 1591) & 2688.36 & (0, 458) \\
        \textit{Last-1 Top-2} & 249.5 & (0, 588) & 273.5 & (0, 1384) & 477.0 & (0, 2100) & 243.73 & (0, 506) & 2609.71 & (0, 2570) \\
        \bottomrule
    \end{tabular}
    }

    \label{tab:magnitude_trend}
\end{table*}

\subsection{Outlier Smoothing for Low-precision LLM Quantization}
\label{sec:bg:llm-quantization}
Reducing quantization error is crucial for achieving effective low-precision model quantization. However, as highlighted by LLM.int8()~\cite{dettmers2022gpt3}, directly quantizing LLMs to INT8 leads to significant accuracy loss due to the presence of outliers. To address these outliers, LLM.int8() employs a mixed-precision decomposition scheme. While this approach preserves model accuracy, its fine-grained decomposition introduces computational overhead and potential performance bottlenecks. 

Olive~\cite{guo2023olive} addresses the impact of MAs on low-precision quantization by proposing a hybrid quantization scheme that quantizes MAs separately from the remaining elements. Similarly, PrefixQuant~\cite{chen2024prefixquant} groups tokens with MAs and jointly quantizes them, resulting in reduced quantization error. This approach has also been applied to KV cache quantization~\cite{zhang2024pyramidkv}, following the same principle. Collectively, these studies highlight the critical importance of understanding outlier behavior within LLMs to develop more effective quantization strategies.

On the other hand, to eliminate the channel-wise outliers, SmoothQuant~\cite{xiao2023smoothquant} proposes migrating the quantization challenge from activations to weights using scale invariance. This allows INT8 quantization for both weights and activations across all matrix multiplications in LLMs. Outlier Suppression+~\cite{wei2023outlier} further enhances quantization by introducing a fast and stable scheme for calculating scaling values, effectively balancing the quantization burden. 

To reduce manual intervention and improve performance under extremely low-bit quantization, OmniQuant~\cite{shao2023omniquant} introduces Learnable Weight Clipping and Learnable Equivalent Transformation, optimizing both weight-only and weight-activation quantization processes. In W4A8 quantization with weight clipping, QQQ~\cite{zhang2024qqq} dynamically manages outliers through adaptive smoothing. Additionally, QServe~\cite{lin2024qserve} introduces SmoothAttention to mitigate accuracy degradation caused by 4-bit KV quantization. Both QQQ and QServe have greatly improved LLM accuracy under W4A8 quantization.

While most previous studies focus on mitigating the impact of channel-wise outliers during the quantization process, this work investigates the root causes of both MAs and COs. We propose some insights to address these outliers by targeting and removing them at their fundamental level.

\section{Empirical Study on Massive Activation}
\label{sec:emperiment-study}
\subsection{Settings}
To investigate the formation of massive activations (MAs), we conduct experiments on various LLMs, including the LLaMA series~\cite{touvron2022llama, touvron2023llama}, GPT-2~\cite{achiam2023gpt}, and Qwen~\cite{yang2024qwen2technicalreport}, using two datasets: WikiText~\cite{wikitext} and C4~\cite{c4_dataset}. Each experiment is averaged over 100 random samples from the dataset. LLM performance is evaluated using the perplexity (PPL) metric.  

Following the definition of MAs from~\cite{sun2024massive}, an activation is considered~\textbf{massive} if its magnitude exceeds 100 and is at least 1,000 times greater than the median activation magnitude.

\begin{figure}[t]
    \centering
    \includegraphics[width=1\linewidth]{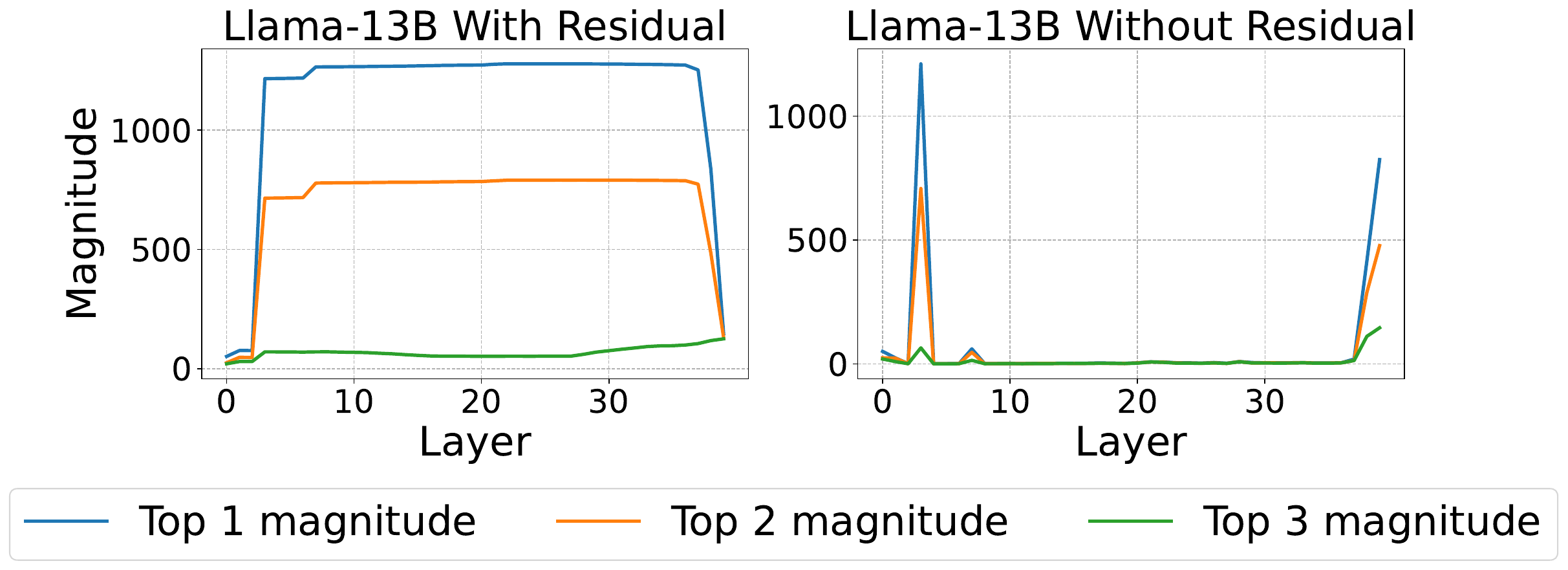}
    \caption{Left: TMAs and FMAs within the input of LLaMA-13B across each layer. Right: after removing the MAs in residual connection, only TMA left.}
    \label{fig:llama-mas}
\end{figure}

\begin{table*}
    \centering
    \caption{Impact of MAs on the performances (in perplexity) of LLaMA, GPT-2, and Qwen models.}
     \resizebox{\textwidth}{!}{
    \begin{tabular}{l>{\centering\arraybackslash}p{1.6cm} >{\centering\arraybackslash}p{1.6cm} | >{\centering\arraybackslash}p{1.6cm} >{\centering\arraybackslash}p{1.6cm} | >{\centering\arraybackslash}p{1.6cm} >{\centering\arraybackslash}p{1.6cm} | >{\centering\arraybackslash}p{1.6cm} >{\centering\arraybackslash}p{1.6cm} | >{\centering\arraybackslash}p{1.6cm} >{\centering\arraybackslash}p{1.6cm}}
        \toprule
        & \multicolumn{2}{c}{\textbf{LLaMA3.2-3B}} & \multicolumn{2}{c}{\textbf{LLaMA3.1-8B}} & \multicolumn{2}{c}{\textbf{LLaMA2-13B}} & \multicolumn{2}{c}{\textbf{GPT-2}} & \multicolumn{2}{c}{\textbf{Qwen2.5-7B}} \\
        \cmidrule(lr){2-3} \cmidrule(lr){4-5} \cmidrule(lr){6-7} \cmidrule(lr){8-9} \cmidrule(lr){10-11}
        \textbf{Intervention} & \textbf{WikiText} & \textbf{C4} & \textbf{WikiText} & \textbf{C4} & \textbf{WikiText} & \textbf{C4} & \textbf{WikiText} & \textbf{C4} & \textbf{WikiText} & \textbf{C4} \\
        \midrule
        \textit{Original} & 5.567 & 10.790 & 6.941 & 9.046 & 4.355 & 6.405 & 14.795 & 19.460 & 6.520 & 11.773 \\
        \textit{TMAs to mean at ${y}_{7}$} & 1124111.75 & 21046.82 & 21281.49 & 1301562.25 & 1301562.25 & 6469.42 & 14.841 & 19.560 & 71216.17 & 66588.86 \\
        \textit{TMAs to zeroes at ${y}_{7}$} & 1138151.23 & 21951.41 & 21601.10 & 1302018.53 & 1309211.61 & 7128.32 & 14.911 & 19.928 & 71835.61 & 67518.35 \\
        \textit{TMAs to mean at ${y}_{6}$} & 6.053 & 14.423 & 7.026 & 10.046 & 4.355 & 6.405 & 14.795 & 19.460 & 6.537 & 11.797 \\
        \textit{TMAs to zeroes at ${y}_{6}$} & 6.237 & 14.767 & 7.147 & 10.255 & 4.371 & 6.498 & 14.831 & 19.565 & 6.642 & 13.021 \\

        \bottomrule
    \end{tabular}
    }
    \label{tab:acc_analysis}
\end{table*}

\subsection{Observations on Massive Activation}
\label{sec:ma-obs}
In our experiments, we investigate the existence of MAs in the hidden state tensors within the attention and MLP blocks. Next, we modify the inference process of LLMs by directly intervening in the layers where massive activations emerge. Specifically, for any hidden state exhibiting massive activations, we manually set those activations to fixed values. The modified hidden state is then passed to the subsequent layer, with the remaining computations proceeding as usual. As a result of these studies, we have the following surprising observations that differ from or were not reported in earlier literature, summarized as follows:

\noindent\textbf{Massive Activations are first appeared in the FFN Block: }We found that for all LLMs, MAs first appear within the feed-forward network (FFN) of first layer. Specifically, in models using gated MLPs, such as the LLaMA series and Qwen, MAs emerge in \( y_{6} \), the product of \( y_{4} \) and \( y_{5} \), as illustrated in Figure~\ref{fig:llm-arch} (c). In contrast, for LLMs with conventional MLPs, like GPT-2, MAs are first produced immediately after the GeLU activation, represented by \( y_{4} \) in Figure~\ref{fig:llm-arch} (d).

\begin{figure}[t]
    \centering
    \includegraphics[width=1\linewidth]{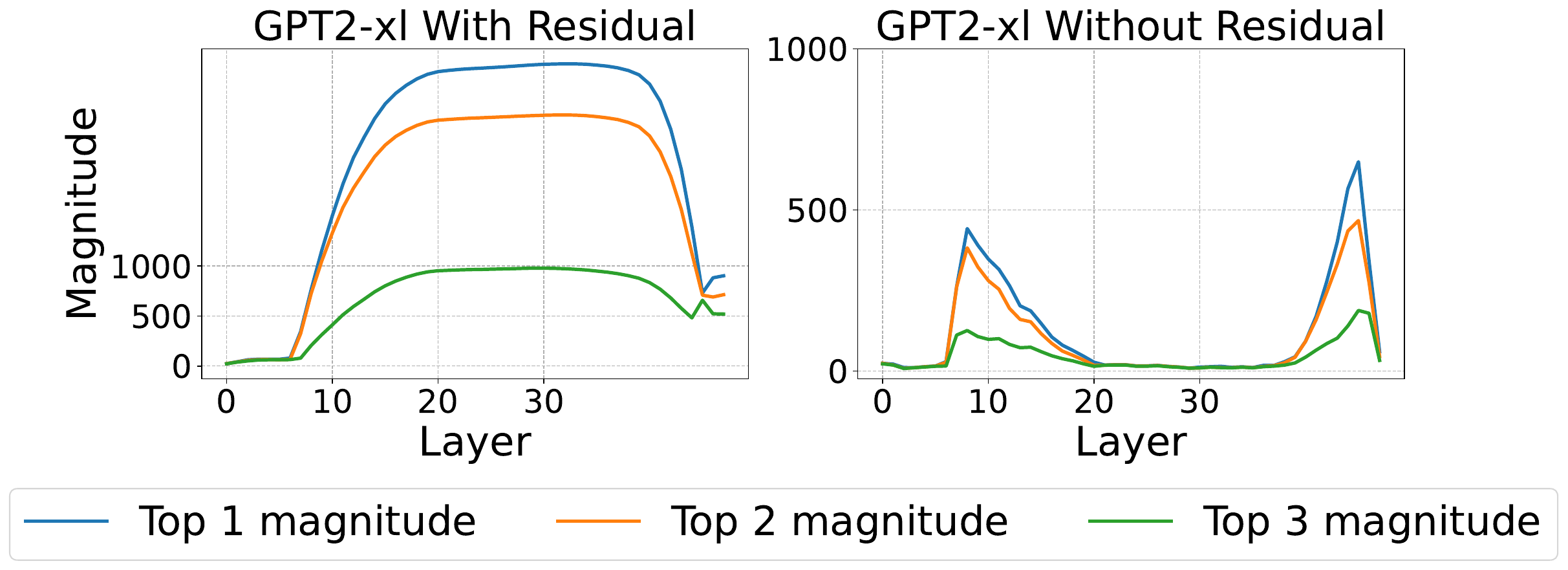}
    \caption{Left: TMAs and FMAs within the input of GPT-2 across each layer. Right: after removing the MAs in residual connection, only TMA left.}
    \label{fig:gpt2-mas}
\end{figure}

\noindent\textbf{Most of MAs are caused by residual connections within LLM: }Among the MAs observed across LLM layers, most are propagated through residual connections in both the self-attention (SA) and FFN blocks. Specifically, after initially appearing in the FFN, the residual links carry these MAs through the inputs of SA and FFN blocks across the middle layers of the LLM. These MAs are not newly generated but are instead carried forward from previously produced MAs through the intermediate layers via residual connections.
For the final layers (e.g., 39th and 40th layers in LLaMA), MAs are generated spontaneously and are not caused by residual connections. To differentiate these MAs, we call the MAs that are caused by the residual link~\textbf{Fake MAs} (FMAs), and rest of MA~\textbf{True MAs} (TMAs).

To illustrate the presence of TMAs and FMAs, we conduct experiments on LLaMA-13B and GPT-2. The left side of Figure~\ref{fig:llama-mas} and Figure~\ref{fig:gpt2-mas} show the top three elements with the highest magnitudes, identified as MAs, across the input of each layer. Building on this, we remove the residual connections for both the SA and FFN layers throughout the entire LLM. The right side of Figure~\ref{fig:llama-mas} and Figure~\ref{fig:gpt2-mas} present the results after these residual connections are removed from all layers. Our observations show that removing the TMAs at ${y}_{6}$ effectively eliminates all TMAs and FMAs. Due to space constraints, we present results only for LLaMA-13B and GPT-2, although similar behaviors are observed in other LLMs.
\\
\begin{figure}
    \centering
    \includegraphics[width=1.0\linewidth]{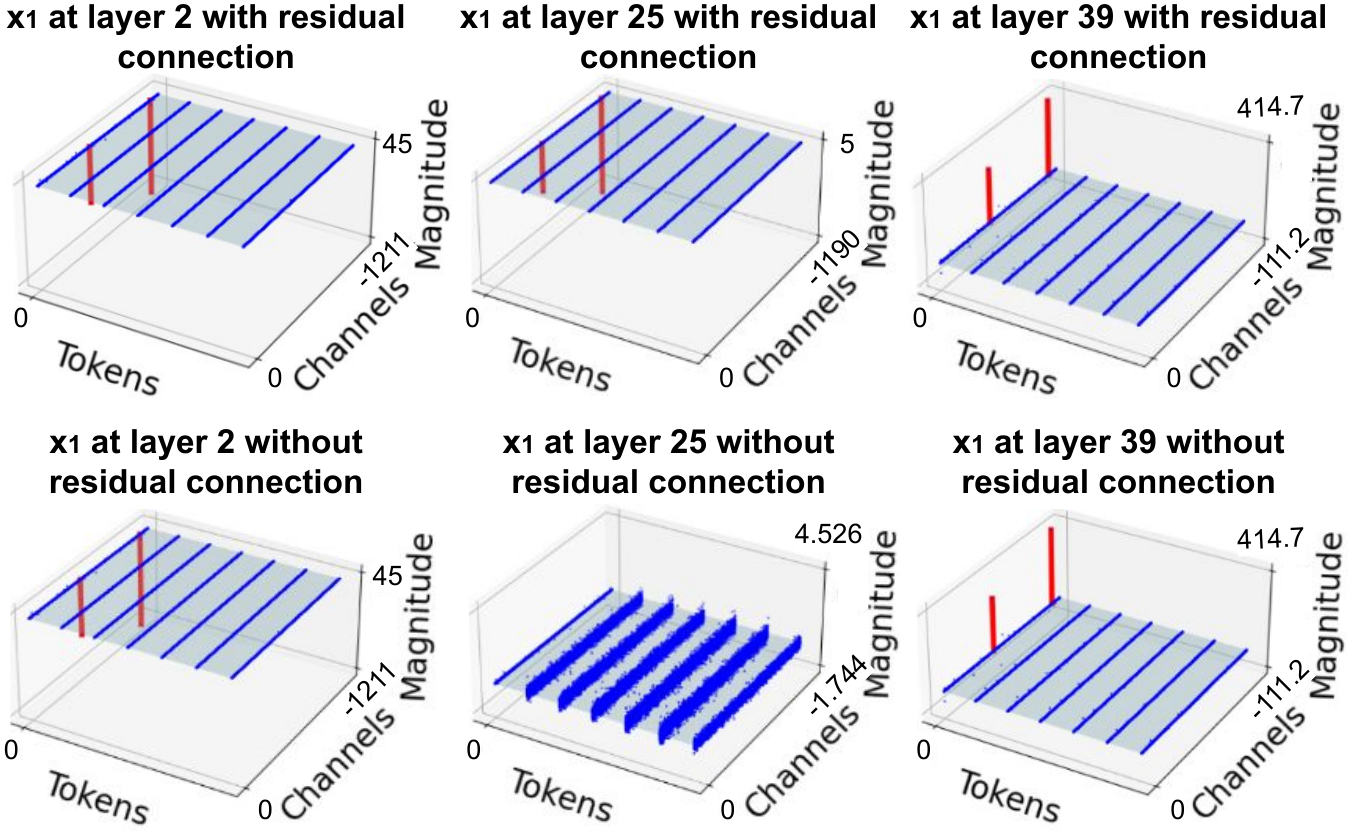}
    \caption{The red lines drawn to the origin plane indicate the MAs. Top three figures are the TMA and FMA of Llama-13B. The bottom three figures are the TMA of Llama 13B model after removing MAs in the residual connection. Layer 4 and Layer 39 have TMA for the same channel and token but with the opposite sign. The MAs of layer 25 is eliminated after the removal of the residual connection.}
    \label{fig:Trends-TMA}
\end{figure}

\noindent\textbf{Trends on TMA Magnitudes: }
Across various models, TMAs exhibit consistent behavior: their magnitude remains fixed within specific channels, regardless of the input sequence tokens. Analyzing the sign of these TMAs reveals a clear pattern: in the final layers, TMAs have a similar magnitude but opposite sign compared to those in the initial layers, occurring at the same channel positions. This indicates that TMAs generated in the early layers are effectively suppressed in the later layers. Table~\ref{tab:magnitude_trend} presents the average magnitudes of TMAs across multiple LLMs, highlighting their presence in the first initial layers and the last two layers. It also shows the top two MAs with the largest magnitudes in each layer's input, along with their corresponding token and channel indices, shown in the first and second number within the bracket. While models like GPT-2 and Qwen display multiple initial and final layers with high activation magnitudes, the observed magnitude and sign trends persist. Figure~\ref{fig:Trends-TMA} shows that layer 2 of LLaMA-13B has a negative TMA while at layer 39 there is a positive TMA at the same channel and token position.



\subsection{Impact of Massive Activation Values on LLM Accuracy}
\label{sec:ma-impacts}
Building on the presence of TMAs, and FMAs, we next analyze their impact on LLM accuracy. Specifically, we replace all TMAs, which are located at $y_{6}$ of FFN with either zero or the mean value of their respective tensors. As shown in Table~\ref{tab:acc_analysis}, the results remain comparable to the original LLM. Notably, for LLaMA2-13B, GPT-2, and Qwen, the PPL values are nearly identical to those of the original LLM on both WikiText-2 and C4, demonstrating that TMAs, and FMAs can be effectively eliminated without any negative impact on accuracy performance.

In contrast, the removal of TMAs located at $y_{7}$ of the FFN results in disastrous effects on LLM performance. As shown in Table~\ref{tab:acc_analysis}, replacing TMAs with mean or zero values significantly increases PPL across models, with the exception of GPT-2. Thus, we show that most TMAs can be safely removed by replacing them with either zero or the mean of the tensor containing them at $y_{6}$. Consequently, no TMAs can appear at Y7 or propagate via the residual connection. More detailed information for MA in MLP and attention blocks is in Appendix \ref{appendix: appendix_A}.

\subsection{Insights for MA Smoothing}
\label{sec:ma-insight}
The presence of MAs is widely acknowledged as a major challenge in LLM quantization, particularly when aiming to enable efficient matrix multiplication within SA. As demonstrated in Section~\ref{sec:ma-obs} and Section~\ref{sec:ma-impacts}, all FMAs can be effectively eliminated by replacing them with either the mean value or zero, with negligible impact on LLM performance. This makes the corresponding activation matrices significantly easier to quantize.  

In contrast, removing TMAs directly leads to severe performance degradation. As a result, existing outlier smoothing techniques, such as mathematical invariance transformations~\cite{ashkboos2024quarot} and~\cite{xiao2023smoothquant}, are typically applied exclusively to these outliers. Since mathematical invariance transformations (e.g., Hadamard transform) introduce additional computational overhead for outlier smoothing, limiting their application to the small number of TMAs significantly reduces the overall computational cost.

\section{Empirical Study on Channel-wise Outliers}
\label{sec:channel-wise-study}

\subsection{Settings}
In addition to the presence of MAs, channel-wise outliers (CO) are also observed within the intermediate results of LLMs, as noted in several prior studies~\cite{xiao2023smoothquant, ashkboos2024quarot, quip, liu2024spinquant}. These outliers significantly degrade the performance of low-precision LLM quantization.
Following our study on MAs, we examine the presence and formation of channel-wise outliers in various LLMs (LLaMA series and GPT-2) using two datasets: WikiText and C4. LLM performance is assessed using the perplexity (PPL) metric, with each experiment averaged over 100 random samples.
Since no formal study has been conducted on channel-wise outlier before, we use the following criteria to search for the channel-wise outlier.

For each channel \( A_j \) within an activation matrix \( A \), it is classified as an~\textbf{outlier channel} if it satisfies the following criteria:
\begin{itemize}
    \item The mean of \(A_j\) exceeds the overall average of the tensor by more than \(m\sigma_{A}\), where \(m\) is a predefined parameter and \(\sigma_{A}\) is the standard deviation of elements within \(A\).
    \item The standard deviation of \(A_j\) is below a threshold \(\beta\).
\end{itemize}

The first criterion ensures that the average value of the entire channel is sufficiently high to qualify as an outlier, while the second criterion ensures that all elements within the channel have similar magnitudes, aligning with outlier channel behavior. Without loss of generality, in the following experiment, \(m\) is set to 4, and \(\beta\) is set to \(1/3\). We also present the results under different settings in the subsequent sections.

\subsection{Observations on channel-wise Outliers}
\label{sec:co-obs}
\begin{figure}
    \includegraphics[width=1\columnwidth]{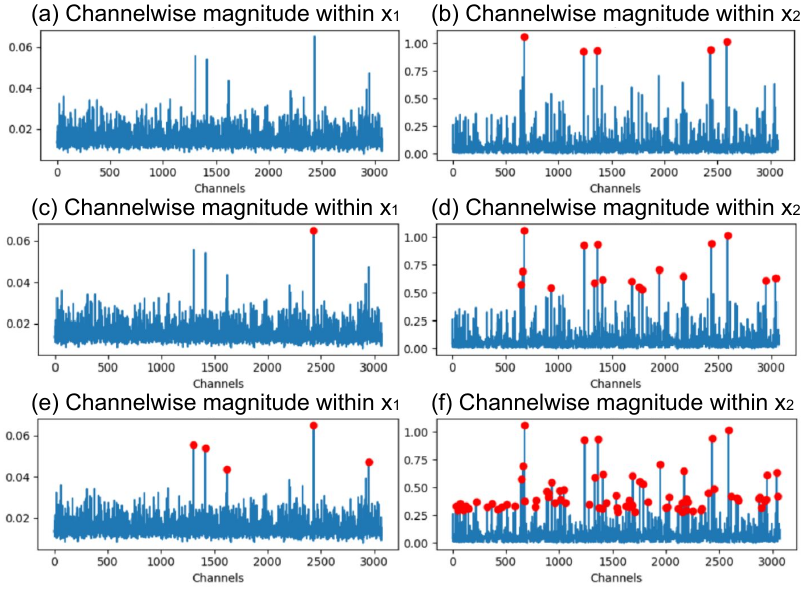}
    \caption{Changes on number of channel-wise outlier after the input $x_{1}$ of the first layer of LLaMA-13B passing through the RMSNorm. Outlier channels denoted by red dots. (a) and (b) shows the results by setting m=6, (c) and (d) for m=4, and (e) and (f) for m=2, respectively.}
    \label{fig:channel-wiseoutlier1}
\end{figure}
\begin{figure*}[t]
    \centering
    \includegraphics[width=2.1\columnwidth]{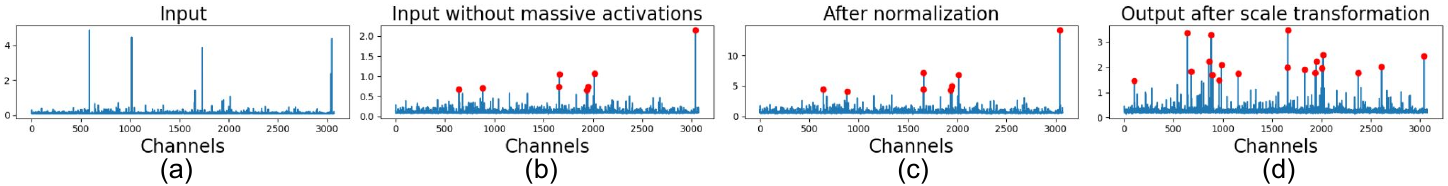}
    \caption{Changes on number of channel-wise outliers within a normalization block of SA. Outlier channels denoted by red dots. The channel-wise means of (a) the input $x_{1}$, (b) $x_{1}$ after removing the MAs, (c) the output of the standardization operation, and (d) the output of normalization $x_{2}$ are plotted. A similar observation has been observed for the normalization block within FFN.}
    \label{fig:channel-wiseoutlier2}
\end{figure*}

We examine the presence of outlier channels in the input, output, and hidden state tensors within the SA and MLP blocks. These correspond to the inputs and outputs of the SA and FFN blocks (e.g., \(x_1\), \(x_2\), \(y_1\), and \(y_2\)) as well as intermediate results (e.g., \(x_3\), \(x_4\), \(y_3\), and \(y_5\)) depicted in Figure~\ref{fig:llm-arch} (b), (c), and (d).
Next, we delve deeper into normalization operation and attention weight matrix multiplications, looking at how the learned model weights associated with each of these transformations affect the occurrence of outlier channels in the output activations. Specifically, we observe the effects of smoothening the outlier channels within these weights by replacing them with fixed values. The observations are summarized below.

\noindent\textbf{Channel-wise outliers first arise after the normalization operation in first layer: }We observe that in all evaluated LLMs, outlier channels first emerge during the initial normalization operation preceding the SA block. Figure~\ref{fig:channel-wiseoutlier1} illustrates the average magnitude of each channel in the input and output of the normalization operation within the first layer of the SA block. Red dots represent the outlier channels. The results are shown by varying the criteria for channel-wise outliers, with \(m\) set to 2, 4, and 6, respectively. Notably, the number of outlier channels increases largely after the normalization operation.

\noindent\textbf{Learned rescaling operations inside Normalization block produces outlier channel:} As shown in Figure~\ref{fig:llm-profile} (d), the normalization operation with in LLM, such as LayerNorm or RMSNorm, are further contains two components: standardization and rescaling. For example, in LayerNorm, the input is first normalized by subtracting its mean $\mu$ and dividing by its standard deviation $\sigma$. Each channel of the normalized output is then scaled by a learnable parameter $\gamma$ and shifted by another learnable parameter $\beta$. 

We conduct an outlier analysis of tensors within the normalization block, as illustrated in Figure~\ref{fig:channel-wiseoutlier2}. To isolate the effects of channel-wise outliers, we first eliminate massive activations (MAs) from the input, allowing for a clearer visualization of outlier channels. In the normalization process, the inputs undergo token-wise standardization followed by a rescaling operation. Our findings reveal that the standardization step does not introduce additional channel-wise outliers (Figure~\ref{fig:channel-wiseoutlier2} (c)). However, the rescaling operation has a channel-specific impact, which can lead to an increase in channel-wise outliers, as depicted in Figure~\ref{fig:channel-wiseoutlier2} (d).

\begin{figure}[t]
    \includegraphics[width=1\columnwidth]{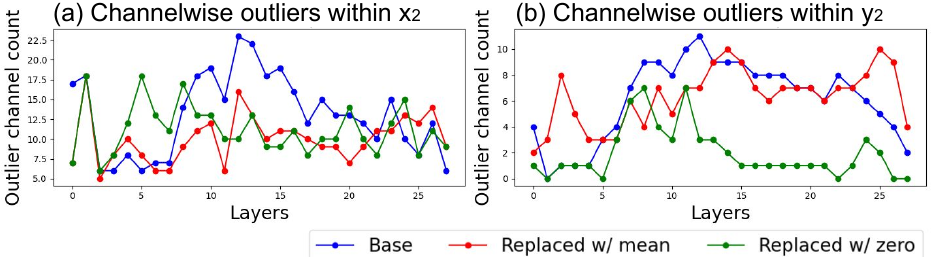}
    \caption{The blue line shows the number of outlier channels in the normalization layer inputs for each LLM layer. To identify the source of these outliers, we examine the corresponding rescaling factors \(\gamma\) that contribute to the channel-wise outliers. These rescaling factors are then replaced with either their mean values (red lines) or zeros (green lines).}
    \label{fig:channel-wiseoutlier4}
\end{figure}

To further validate the impact of the rescaling operation, we modify the rescaling factor vector \(\gamma\) by identifying the indices associated with the outlier channels in the output of the normalization operation. This modification was applied to the normalization layers within both SA and FFN layers. Specifically, the rescaling factor elements at these indices were replaced with either the mean of the rescaling vector or zero. Both modifications result in a noticeable reduction in the number of outlier channels in the subsequent outputs, as shown in Figure~\ref{fig:channel-wiseoutlier4}.

\subsection{Observations on channel-wise Outliers in weight matrix multiplications}
\label{sec:co-weight}
In this section, we examine the presence of channel-wise outliers during matrix multiplication with weight tensors. As a case study, we focus on the Query weight matrix (\(\mathbf{W_q}\)) within the SA block, and Key and Value matrices have the same trends. When examining the output activations ($\mathbf{x_3}$), new channel-wise outliers emerge that are absent in the input activations ($\mathbf{x_2}$). Specifically, $\mathbf{x_3}$ can be computed as follows:
\begin{equation}
\mathbf{x_3} = \mathbf{W_q} \cdot \mathbf{x_2} 
\end{equation}

If channel-wise outliers are observed in $\mathbf{x_3}$ but not in the corresponding input activations channel $\mathbf{x_2}$, we hypothesize that specific channels (rows) in $\mathbf{W_q}$ are responsible for the existence of new channel-wise outlier. These channels, which constitute approximately $1\%$ of all channels within $W_{q}$, appear to hold greater numerical importance compared to others on LLM accuracy. We call it~\textit{Outlier Triggering Channels} (OTC).

An important but subtle observation is that these OTCs do not exhibit outlier characteristics when analyzing \(\mathbf{W_q}\) alone, based on mean and standard deviation statistics. However, their interaction with \(\mathbf{x_2}\) gives rise to outlier activations in \(\mathbf{x_3}\). This finding highlights the critical role of OTCs in outlier formation, despite their seemingly unremarkable statistical profile in isolation. To test this hypothesis, we evaluate model performance by modifying \(\mathbf{W_q}\) in two ways: (a) setting all elements within the OTCs, which comprise approximately $1\%$ of the total number of channels in \(\mathbf{W_q}\), to their mean values, and (b) setting a random $1\%$ of channels to their mean values. The results of these interventions are presented in Table~\ref{tab:intervention_analysis}. The modification on OTC will cause a greater accuracy drop than that on equivalent amount of random channels. This comparison highlights the importance of specific weight channels that contribute to the presence of channel-wise outliers on the LLM accuracy. Similar studies have been performed on the key matrix and observe the trend being similar to query matrix, while the value matrix does not follow this trend and remain unaffected even after removing the OTC. 

\subsection{Insights for channel-wise Outlier Smoothing}
\label{sec:co-insight}
Based on the results presented in Section~\ref{sec:co-obs} and Section~\ref{sec:co-weight}, we conclude that the rescaling factor \(\gamma\) in the rescaling operations within the normalization layer plays a significant role in determining the number of channel-wise outliers in \(x_2\) and $y_2$. These outliers are subsequently propagated into the matrix multiplication processes. To effectively mitigate channel-wise outliers in the input, a great strategy is to fine-tune the rescaling factors \(\gamma\) to reduce their variation. This adjustment results in \(x_2\) having fewer outlier channels. However, simply setting the corresponding rescaling factor to a fix value will lead to significant accuracy drop.

OTCs within the weight matrices greatly contribute to channel-wise outliers in the intermediate results of LLMs. A potential solution to address this issue is to adopt parameter-efficient fine-tuning techniques, which can effectively eliminate OTCs without requiring extensive changes to the model.

\begin{table}[t]
    \centering
    \caption{Analysis on the Importance of OTC, other LLMs also have similar trends}
    \resizebox{\columnwidth}{!}{
    \begin{tabular}{l>{\centering\arraybackslash}p{1.3cm} >{\centering\arraybackslash}p{1.3cm} | >{\centering\arraybackslash}p{1.3cm} >{\centering\arraybackslash}p{1.3cm} | >{\centering\arraybackslash}p{1.3cm} >{\centering\arraybackslash}p{1.3cm}}
        \toprule
        & \multicolumn{2}{c}{\textbf{LLaMA3.2-3B}} & \multicolumn{2}{c}{\textbf{LLaMA3.1-8B}} & \multicolumn{2}{c}{\textbf{LLaMA2-13B}} \\
        \cmidrule(lr){2-3} \cmidrule(lr){4-5} \cmidrule(lr){6-7}
        \textbf{Intervention} & \textbf{WikiText} & \textbf{C4} & \textbf{WikiText} & \textbf{C4} & \textbf{WikiText} & \textbf{C4} \\
        \midrule
        \textit{base model} & 5.567 & 10.790 & 6.941 & 9.046 & 4.355 & 6.405 \\
        \textit{Remove OTCs} & 38.924 & 165.396 & 480.8123 & 465.2235 & 774.7298 & 15398.1279 \\
        \textit{Remove random channels} & 7.5094 & 11.990 & 7.1700 & 18.602 & 4.4455 & 6.682 \\
        \bottomrule
    \end{tabular}
    }
    \label{tab:intervention_analysis}
\end{table}


\section{Conclusion}
Outliers in LLMs are crucial to address because of their significant impact on the accuracy of quantized LLMs. In this paper, we undertake a detailed investigation into the mechanisms behind the formation of outliers and develop strategies to mitigate their effects. We explore the causes of these outliers and propose practical approaches for their elimination, setting the stage for more efficient quantization processes. 

Our comprehensive analysis not only highlights the challenges posed by outliers but also provides innovative solutions that could be pivotal for the advancement of quantization techniques in LLMs. We hope our findings make a valuable contribution to the ongoing research within the LLM community, especially in addressing the complexities of quantization challenges presented by outliers.

\newpage
\section*{Limitations}
While this survey offers a comprehensive overview of outliers within LLMs, it is important to acknowledge some limitations. The study of outliers is specifically tailored to LLMs, and there is scope for extending this research to other types of large models that handle multimodal inputs. Further investigation in these areas could provide a broader understanding of outlier effects across different model architectures and enhance the robustness of multimodal systems.


\appendix

\section{Activation Statistics Across All Sublayers}
\label{appendix: appendix_A}
To support the claim that TMAs occur only at y6 (and y7 if un-smoothed), we added Table \ref{tab:mlp-ma-removal} and Table \ref{tab:attention-layer2}, listing the top-2 absolute activation values at every sublayer ($x_{1}$--$x_{9}$, $y_{1}$--$y_{7}$) of layer 2 under a single WikiText input, both with and without MA removal at Y6. 
\begin{itemize}
  \item There are no MA observed from $x_{1}$--$x_{9}$.
  \item From  $y_{1}$--$y_{5}$the top two values remain unchanged by MA smoothing.
  \item At y6, the maximum absolute value drops from $-499.25$ to $37.09$, and at Y7 from $328.25$ to $24.125$.
\end{itemize}

\begin{table}[h]
  \centering
  \caption{Top-2 activations in MLP $y_{1}$--$y_{7}$ of LLaMA-3.2-3B layer 2, with and without MA removal at Y6.}
  \small
  \label{tab:mlp-ma-removal}
  \begin{tabular}{lcc}
    \toprule
    Sublayer & With MA & Without MA (y6 only) \\
    \midrule
    $y_1$ & $-16.03$, $10.01$ & $-16.03$, $10.01$ \\
    $y_2$ & $4.19$, $3.87$ & $4.19$, $3.87$ \\
    $y_3$ & $13.67$, $8.5$ & $13.67$, $8.5$ \\
    $y_4$ & $-36.5$, $-8.92$ & $-36.5$, $-8.92$ \\
    $y_5$ & $13.67$, $8.5$ & $13.67$, $8.5$ \\
    $y_6$ & $-499.25$, $-37.09$ & $-37.09$, $5.09$ \\
    $y_7$ & $328.25$, $-303.25$ & $24.125$, $-22.375$ \\
    \bottomrule
  \end{tabular}
\end{table}

\begin{table}[h]
  \centering
  \caption{Top-2 absolute values from Attention $x_{1}$--$x_{7}$ in layer 2 of LLaMA-3.2-3B.}
  \label{tab:attention-layer2}
  \begin{tabular}{lc}
    \toprule
    Sublayer & Top-2 Values \\
    \midrule
    $x_1$ & $-15.93$, $10.58$ \\
    $x_2$ & $3.48$, $3.47$ \\
    $x_3$ & $-8.16$, $-8.05$ \\
    $x_4$ & $-9.94$, $-9.78$ \\
    $x_5$ & $-0.73$, $0.71$ \\
    $x_6$ & $-14.77$, $-14.51$ \\
    $x_7$ & $0.99$, $0.98$ \\
    $x_8$ & $0.47$, $0.45$ \\
    $x_9$ & $-0.36$, $-0.34$ \\
    \bottomrule
  \end{tabular}
\end{table}

\section{Outlier-Channel Ablation Study}
In Table \ref{tab:channel_replacement_effect}, we report the perplexity of LLaMA-3.2-3B on Wikitext-2 when replacing channels beyond $\{6,4,2\}$ standard deviations (SD) in QKV, LayerNorm rescaling factor and MLP weights with (a) the channel mean, and (b) random channel replacements. 

\begin{itemize}
  \item True outlier removals increase PPL from 7.83 to as high as 61.94.
  \item Outliers in LayerNorm rescaling factors have a pronounced impact on perplexity, suggesting their critical role in maintaining performance.
  \item Replacing an equal number of randomly selected channels results in considerably smaller degradation in PPL for most cases.
  \item Interestingly, for QKV projections, a more aggressive outlier threshold (2SD) results in lower perplexity (10.97) compared to replacing an equivalent number of random channels (14.04), indicating that these outliers may be less essential to model performance.
\end{itemize}

  

\begin{table}[t]
    \centering
    \caption{Table A.3: Wikitext-2 PPL under outlier vs.\ random channel replacements at different thresholds for LLaMA-3.2-3B. The base model perplexity is 7.8316.}
    \resizebox{\columnwidth}{!}{
    \begin{tabular}{l >{\centering\arraybackslash}p{1.8cm} >{\centering\arraybackslash}p{1.8cm} >{\centering\arraybackslash}p{1.8cm}}
        \toprule
        \textbf{Intervention (setting to mean)} & \textbf{6 SD} & \textbf{4 SD} & \textbf{2 SD} \\
        \midrule
        \textit{QKV outliers} & 7.8395 & 7.9201 & 10.9746 \\
        \textit{QKV random} & 7.8315 & 7.8419 & 14.0361 \\
        \textit{LayerNorm outliers} & 8.3497 & 11.4209 & 61.9459 \\
        \textit{LayerNorm random} & 7.8338 & 7.9110 & 8.0399 \\
        \textit{MLP outliers} & 7.8327 & 7.9108 & 15.3786 \\
        \textit{MLP random} & 7.8322 & 7.8452 & 9.5039 \\
        \bottomrule
    \end{tabular}
    }
    \label{tab:channel_replacement_effect}
\end{table}

\end{document}